\documentclass{article}

\usepackage[preprint]{nips_2018}

\usepackage[utf8]{inputenc} 
\usepackage[T1]{fontenc}    
\usepackage{hyperref}       
\usepackage{url}            
\usepackage{booktabs}       
\usepackage{amsfonts}       
\usepackage{nicefrac}       
\usepackage{microtype}      
\usepackage{graphicx}
\usepackage{subcaption}

\title{Deep Convolutional Neural Network for Plant Seedlings Classification}

\author{
  Daniel Nkemelu, Daniel Omeiza, Nancy Lubalo \\
  \texttt{[dnkemelu, domeiza, nlubalo]@africa.cmu.edu} \\
}

\begin{document}

\maketitle

\begin{abstract}
  Agriculture is vital for human survival and remains a major driver of several economies around the world; more so in underdeveloped and developing economies. With increasing demand for food and cash crops, due to a growing global population and the challenges posed by climate change, there is a pressing need to increase farm outputs while incurring minimal costs. Previous machine vision technologies developed for selective weeding have faced the challenge of reliable and accurate weed detection. We present approaches for plant seedlings classification with a dataset that contains 4,275 images of approximately 960 unique plants belonging to 12 species at several growth stages. We compare the performances of two traditional algorithms and a Convolutional Neural Network (CNN), a deep learning technique widely applied to image recognition, for this task. Our findings show that CNN-driven seedling classification applications when used in farming automation has the potential to optimize crop yield and improve productivity and efficiency when designed appropriately.
\end{abstract}

\section{Introduction}
Plants continue to serve as a source of food and oxygen for all life on earth. In continents like Africa, where agriculture is predominant, proper automation of the farming process would help optimize crop yield and ensure continuous productivity and sustainability . The transformation of the agricultural sector by use of smart farming methods can power economic growth in many countries. According to \cite{fao}, there is a strong link between increased productivity and economic prosperity.

One major reason for reduction in crop yield is weed invasion on farmlands. Weeds generally have no useful value in terms of food, nutrition or medicine yet they have accelerated growth and parasitically compete with actual crops for nutrients and space. Inefficient processes such as hand weeding has led to significant losses and increasing costs due to manual labour \cite{gharde}. Precision agriculture, with the goal of defining systems that support decision-making in farm management in order to optimize returns on outputs while preserving resources, and weed control systems have been developed aiming at optimizing yields and costs while minimizing environmental challenges; some robotic systems have been used to do this \cite{dyrmann}. The robots and the vision machines need to be able to precisely and reliably detect a weed from the useful plants. Machine vision technologies developed for selective weeding face a challenge of reliable and accurate weed detection. It’s not easy to identify the weeds due to unclear crops boundaries, with varying rocky or sandy backgrounds, and as a result, traditional classification methods are likely to fail on this task \cite{lee}.

In this work, we explore the performance of traditional computer vision methods on this task and show that a Deep Convolutional Neural Network (CNN) does the best job at classifying plant seedlings. In computer vision, CNNs have been known to be powerful visual models that yield hierarchies of features enabling accurate segmentation. They are also known to perform predictions relatively faster than other algorithms while maintaining competitive performance at the same time \cite{shang}.  

\subsection{Data}

The dataset used, provided by the Aarhus University Signal Processing group, in collaboration with University of Southern Denmark, contains a set of 4275 images of approximately 960 unique plants belonging to 12 species at several growth stages. The data is primarily targeted for research that tries to identify plant species at their early growth stage. This allows farmers to conduct weeding before the weeds start competing with crops for nutrition. Performing image segmentation at this stage is also easier because there is less over-lapping of leaves \cite{gisselson}.

\begin{figure}
  \centering
  \includegraphics[width=0.6\linewidth]{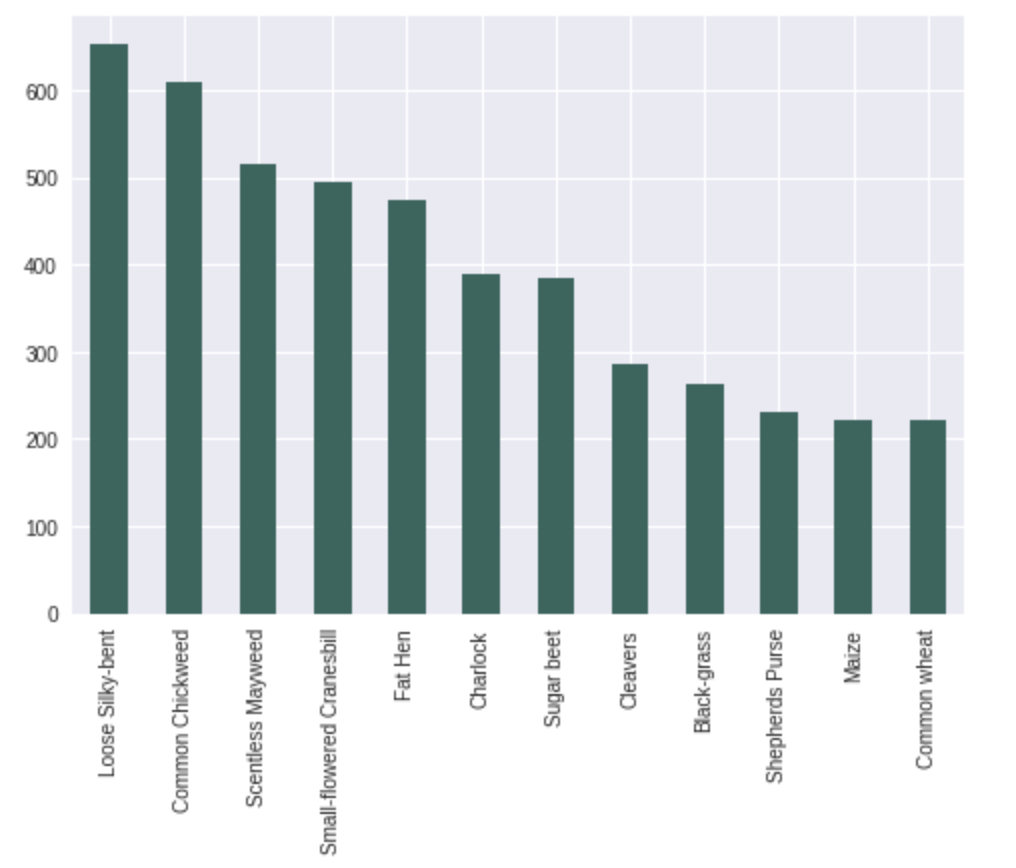}
  \caption{Bar graph showing the distribution of the different classes of plants}
  \label{fig:dist_plant}
\end{figure}

\begin{figure}
  \centering
  \includegraphics[width=0.6\linewidth]{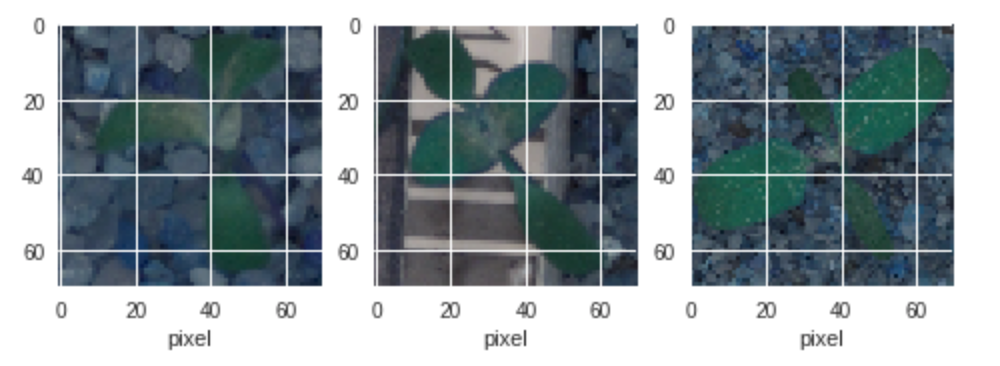}
  \caption{Random samples of images in the dataset}
  \label{fig:sample_plant}
\end{figure}

\section{Related Work}
\label{gen_inst}

\cite{gisselson} in their paper provided a dataset that is aimed at ground-based weed or specie spotting and also suggested a benchmark measure to researchers to enable easy comparison of classification results. \cite{han} previously demonstrated the effectiveness of a convolutional neural network to learn unsupervised feature representations for 44 different plant species with high accuracy.

In the course of exploring the right architecture for our model, we consider the work of \cite{wangsu} in classifying leaves using the GoogLeNet and VGGNet architectures. \cite{sun} implemented a 26-layer deep learning model consisting of 8 residual blocks in their classification of 10,000 images of 100 ornamental plant species achieving classification rates of up to 91.78\%.

\cite{milioto} addressed the problem of CNN-based semantic segmentation of crop fields separating sugar beet plants, weeds, and background solely based on RGB data by proposing a deep encoder-decoder CNN for semantic segmentation that is fed with a 14-channel image storing vegetation indexes and other information that in the past has been used to solve crop-weed classification.

\section{Baseline: Using traditional algorithms}
\label{headings}
\paragraph{Data Processing}
We performed image preprocessing on the dataset before training the models. First using Guassian Blur, we smoothen the image, removing high frequency content and then converting this blurred version to HSV space. We created a mask by specifying a range of possible color values of the seedlings to be captured and using a morphological erosion with an 11x11 structuring kernel, we are able to produce foreground seedling images with the backgrounds subtracted.

Below are images of a subset of the seedlings before and after background subtraction. 

\begin{figure}[h!]
  \centering
  \begin{subfigure}[b]{0.55\linewidth}
    \includegraphics[width=\linewidth]{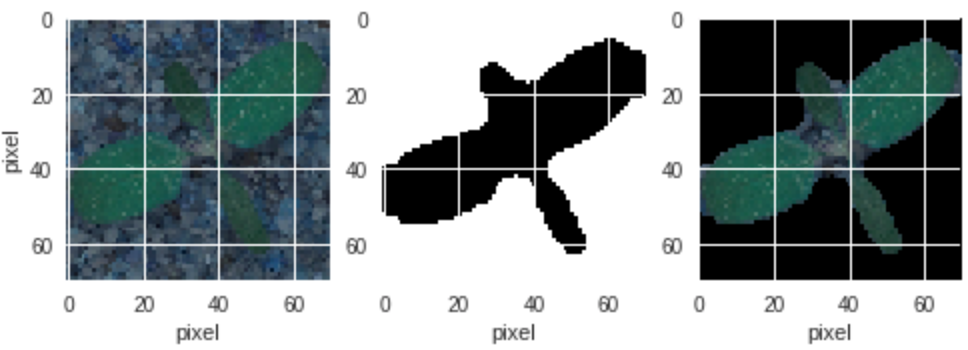}
    \caption{Background segmentation result}
  \end{subfigure}
  \begin{subfigure}[b]{0.55\linewidth}
    \includegraphics[width=\linewidth]{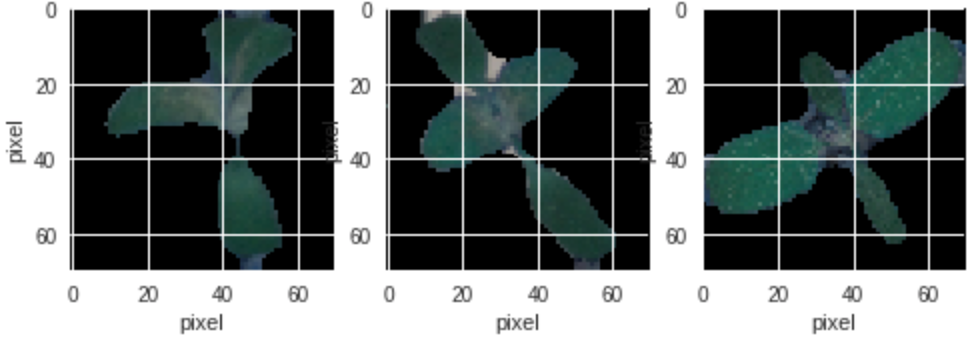}
    \caption{Results of the samples in figure 2 with segmented background}
  \end{subfigure}
  \caption{Images of the segmentation process and a segmented copy of Figure 1.}
  \label{fig:show_seg}
\end{figure}

We then perform image normalization by subtracting the mean from
each pixel, dividing by the standard deviation and then scaled the data in the range of [0,1].

\paragraph{Support Vector Machines and K-Nearest Neighbours}
To perform baseline tests, we use non-neural network techniques: Support Vector Machines (SVMs) and K-Nearest Neighbours classifiers. To find optimal parameters for each model, we perform a grid search using a combination of parameters. For the KNN model, using values for the number of neighbours parameter, $\textit{n\_neighbours}$, ranging from [1, $\sqrt{n}]$, where n = input size, we use grid search to find our best model to have, number of neighbours = 5 with uniform weights and accuracy of 56.84\%.

For our SVM classifier, using similar grid search technique, we find our optimal parameters to be: penalty parameter of the error term, $\textit{C=5}$, kernel=$\textit{linear}$ and gamma value=$\textit{auto}$ which uses $\textit{1 / n\_features}$ as the kernel coefficient. Our accuracy for this model is 61.2\%.

\section{Convolutional Neural Network (CNN): Results and Comparisons}
We attempt to use a CNN for this problem. CNNs have been widely used for diverse image classification tasks. We use two types of input sets; a first case where we pass in the original image pixels and a second case where we performed OpenCV preprocessing of the input image data as in the baseline. 

The neural network architecture has 6 convolutional layers. Each is followed with a rectified linear unit (ReLU). The first two convolutional layers have 64 filters, the next has 128 while the last one has 256. Each convolutional layer has zero padding. After each pair of convolutional layer, we have a max pooling layer for dimensionality reduction and a 10\% dropout to prevent over-fitting. At the end of the six convolutional layers are 3 fully connected layers. The last fully connected layer has a softmax activation function which outputs probability distribution for each of the 12 classes. We use Adam optimizer with a batch size of 32 for each step and a weighted cross-entropy loss, to handle the imbalanced number of pixels for each class. 

For our model using CNNs with attention and no background subtraction using OpenCV, we obtain training accuracy of 98.9\% and validation accuracy of 80.21\%. When we preprocess the data using OpenCV before training the CNN model, we obtain 98.3\% accuracy on training data and 92.6\% on validation data. We see that using traditional image segmentation techniques and CNNs, we are able to increase our accuracy by over 12\% compared to using just CNNs with attention.

\begin{figure}
  \centering
  \includegraphics[width=0.5\linewidth]{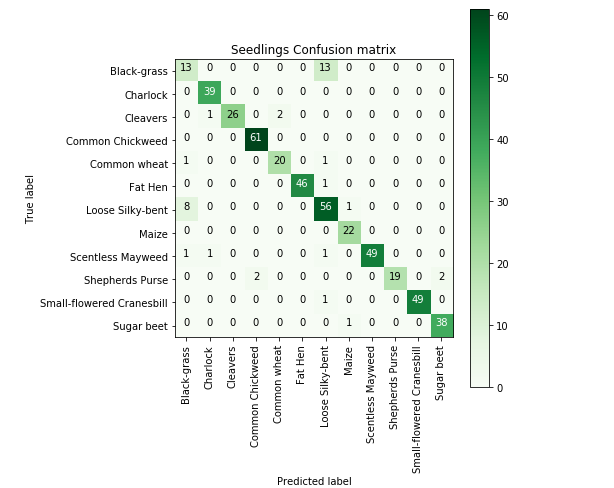}
  \caption{Confusion matrix showing correct classifications in the diagonal and mis-classifications otherwise.}
  \label{fig:conf_mat}
\end{figure}
\begin{table}
  \caption{Table showing performance of the algorithms}
  \label{sample-table}
  \centering
  \begin{tabular}{llll}
    \toprule
    Algorithm     & Description     & Accuracy (\%) \\
    \midrule
    KNN     & with OpenCV background segmentation      & 56.84 \\
    SVM     & with OpenCV background segmentation      & 61.47   \\
    CNN     & with attention   & 80.21   \\
    CNN     & with OpenCV background segmentation      & 92.60 \\
    \bottomrule
  \end{tabular}
\end{table}

\section{Conclusion and Future Work}
\label{others}
An efficient deep learning model for seedlings classification can help farmers optimize crop yields and significantly reduce losses. In this paper, we proposed a deep convolutional neural network method for plant seedlings classification. A dataset that contains images of approximately 960 unique plants belonging to 12 species at several growth stages was used. The model can detect and differentiate a weed from other plants in the wild. A baseline version of the proposed system achieves an accuracy of approximately 93\%. The proposed system can be extended to work with robotic arms for performing actual weeding operation in large farmlands.

For the future of our work, we envision extensions in 3 main directions:
\begin{enumerate}
   \item Training a model with a more inclusive dataset. For instance, using plant seedlings that are more prevalent in African agriculture or other parts of the underdeveloped/developing world other than that of Danish agriculture as provided in our dataset.
   \item Testing out the model using images with multiple plants in a scene. Although the advantage of weeding during plant seedlings early stage is to minimize the challenges that come with overlapping, it would be insightful to see how well the model identifies different classes of plants and potentially predicting the ratio of the classes present.
   \item The long term goal is to be able to embed a more generalized model in a robot arm manipulator which would be used in performing precision agriculture in large farmlands.
\end{enumerate}

We believe that with promising results in classifying plant seedlings, we will be able to completely automate the process of weed control in large farms and thereby reducing costs and manual labour while improving crop yield and productivity.

\end{document}